\documentclass{article}
\usepackage{ijcai20}

\usepackage{times}
\usepackage{soul}
\usepackage{bm}
\usepackage{url}
\usepackage{xcolor}
\definecolor{mydarkgreen}{rgb}{0,0.6,0}
\definecolor{mydarkblue}{rgb}{0,0,0.6}
\usepackage[colorlinks,
linkcolor=black,
citecolor=black,
urlcolor=black]{hyperref}
\usepackage[final]{pdfpages}
\usepackage[utf8]{inputenc}
\usepackage[small]{caption}
\usepackage{graphicx}
\usepackage{subfigure}
\usepackage{amsmath,amssymb,amsfonts}
\usepackage{amsthm}
\usepackage{mathrsfs}
\usepackage{booktabs}
\usepackage{algorithm}
\usepackage{algorithmic}
\usepackage{multirow}
\newcommand{\httpsurl}[1]{\href{https://#1}{\nolinkurl{#1}}}

\newtheorem{theorem}{Theorem}
\newtheorem{definition}{Definition}
\newtheorem{definition*}{Problem}
\newtheorem{remark}{Remark}

\title{Clarinet: A One-step Approach Towards \\ Budget-friendly Unsupervised Domain Adaptation}

\author{
 Yiyang Zhang$^{1,2,}$\footnote{Equal Contribution. See the \emph{clarification letter} on the last page.}
 \and
 Feng Liu$^{2,*}$\and
 Zhen Fang$^{2,*}$\and
 Bo Yuan$^1$\and
 Guangquan Zhang$^2$\And
 Jie Lu$^{2, }$\footnote{Corresponding Author}
 \affiliations
 $^1$Shenzhen International Graduate School, Tsinghua University\\
 $^2$Centre for Artificial Intelligence, University of Technology Sydney\\
 \emails
 zhangyiy18@mails.tsinghua.edu.cn, 
 feng.liu@uts.edu.au,
 zhen.fang@student.uts.edu.au,\\ yuanb@sz.tsinghua.edu.cn,
 \{guangquan.zhang, jie.lu\}@uts.edu.au
 }

\begin{document}

\maketitle

\begin{abstract}
In \emph{unsupervised domain adaptation} (UDA), classifiers for the {target domain} are trained with massive \emph{true-label} data from the {source domain} and unlabeled data from the target domain. However, it may be difficult to collect fully-true-label data in a source domain given limited budget. To mitigate this problem, we consider a novel problem setting where the classifier for the target domain has to be trained with \emph{complementary-label} data from the source domain and unlabeled data from the target domain named \emph{budget-friendly UDA} (BFUDA). The key benefit is that it is much less costly to collect \emph{complementary-label} source data (required by BFUDA) than collecting the \emph{true-label} source data (required by ordinary UDA). To this end, \emph{\underline{c}omplementary \underline{l}abel advers\underline{ari}al \underline{net}work} (CLARINET) is proposed to solve the BFUDA problem. CLARINET maintains two deep networks simultaneously, where one focuses on classifying \emph{complementary-label} source data and the other takes care of the source-to-target distributional adaptation. Experiments show that CLARINET significantly outperforms a series of competent baselines.  
\end{abstract}


\section{Introduction}

\emph{Domain Adaptation} (DA) aims to train a target-domain classifier with data in source and target domains \cite{yan2017learning,Zhou2019TPAMI}. Based on the availability of data in the target domain (e.g., fully-labeled, partially-labeled and unlabeled), DA is divided into three categories: supervised DA \cite{sukhija2016}, semi-supervised DA \cite{ao2017fast,ZHOU2019AIJ} and \emph{unsupervised DA} (UDA) \cite{Gong2018CGDAN,Liu2017,saito2017asymmetric,DBLP:conf/ijcnn/Fang00019}. In practice, UDA methods have been applied to many real-world problems, such as object recognition \cite{DBLP,deng2019cluster,zhao2019geometry}.

UDA methods train a target-domain classifier with massive true-label data from the source domain (true-label source data) and unlabeled data from the target domain (unlabeled target data). Existing works in the literature can be roughly categorised into the following three groups: integral-probability-metrics based UDA \cite{long2015learning}; adversarial-training based UDA \cite{ganin2016domain,long2018conditional}; and causality-based UDA \cite{Gong2016CTC,Gong2018CGDAN,Zhang15MSDA_causal}. Since adversarial-training based UDA methods extract better domain-invariant representations via deep networks, they usually have good target-domain accuracy \cite{sankaranarayanan2018generate}.

\begin{figure}[!tp]
    \centering
    \includegraphics[width=0.35\textwidth]{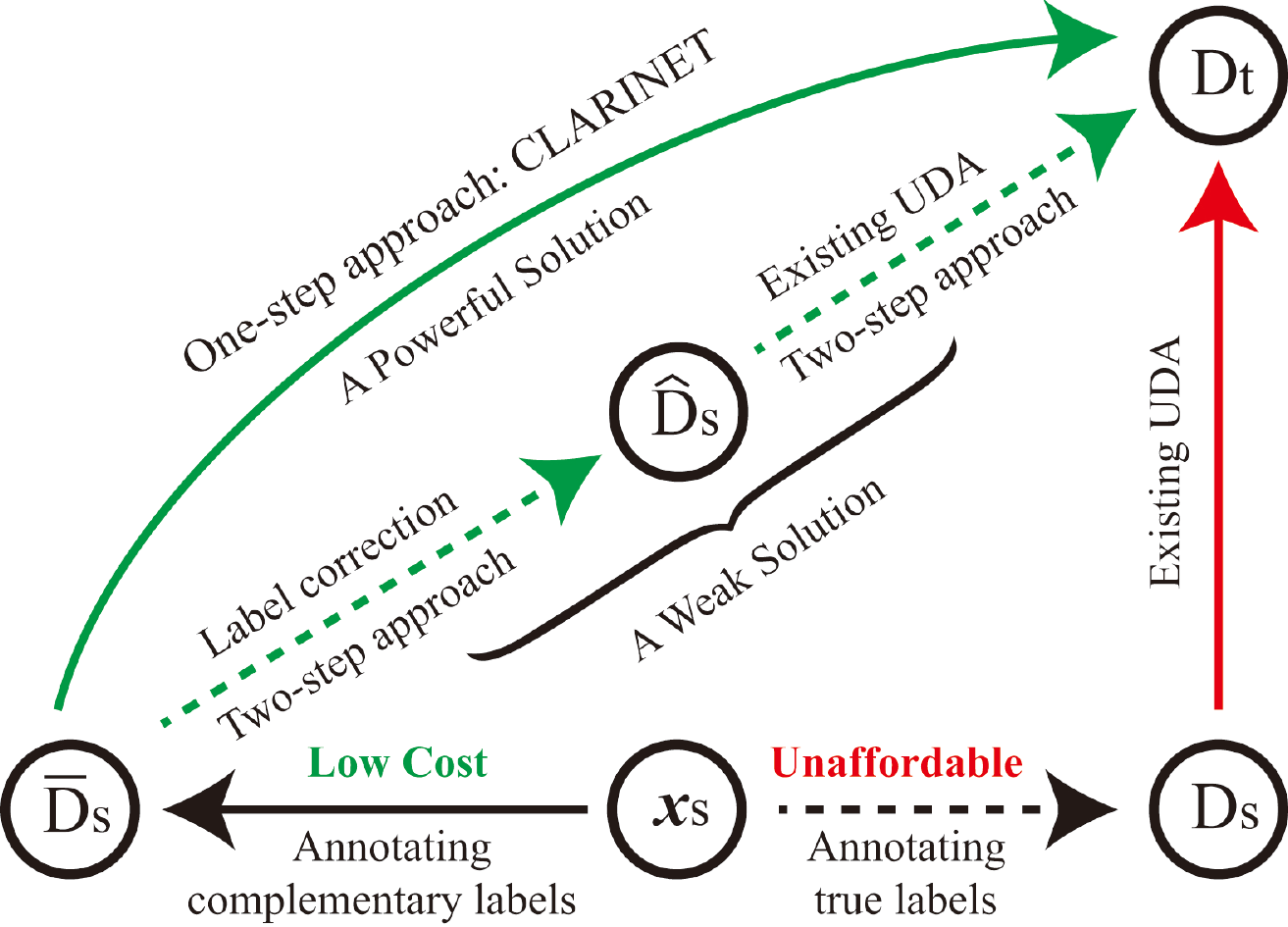}
	\caption{\footnotesize Budget-friendly unsupervised domain adaptation. The red line denotes that UDA methods transfer knowledge from $D_s$ (true-label source data) to $D_t$ (unlabeled target data). However, acquiring fully-true-label source data is \emph{costly} and \emph{unaffordable} (black dash line, $\mathbf{x}_s \rightarrow {D}_s$, $\mathbf{x}_s$ means unlabeled source data). This brings \emph{budget-friendly unsupervised domain adaptation} (BFUDA), namely transferring knowledge from $\overline{D}_s$ (complementary-label source data) to $D_t$. It is much less costly to collect complementary-label source data (black line, required by BFUDA) than collecting the true-label one (black dash line, required by UDA). To handle BFUDA, a weak solution is a two-step approach (green dash line), which sequentially combines complementary-label learning methods ($\overline{D}_s \rightarrow \hat{D}_s$, label correction) and existing UDA methods ($\hat{D}_s \rightarrow D_t$). This paper proposes a one-step approach called \emph{\underline{c}omplementary \underline{l}abel advers\underline{ari}al \underline{net}work} (CLARINET, green line, $\overline{D}_s \rightarrow D_t$ directly).}
	\label{fig: our_solution}
\end{figure}

However, the success of UDA still highly relies on the scale of true-label source data. Namely, the target-domain accuracy of a UDA method (e.g., CDAN) decays when the scale of true-label source data decreases. Hence, massive true-label source data are inevitably required by UDA methods, which is very expensive and prohibitive especially when the budget is limited \cite{liu2019butterfly}. This circumstance may hinder the promotion of DA to more areas.

\begin{figure}[!tp]
	\begin{center}
		{\includegraphics[width=0.48\textwidth]{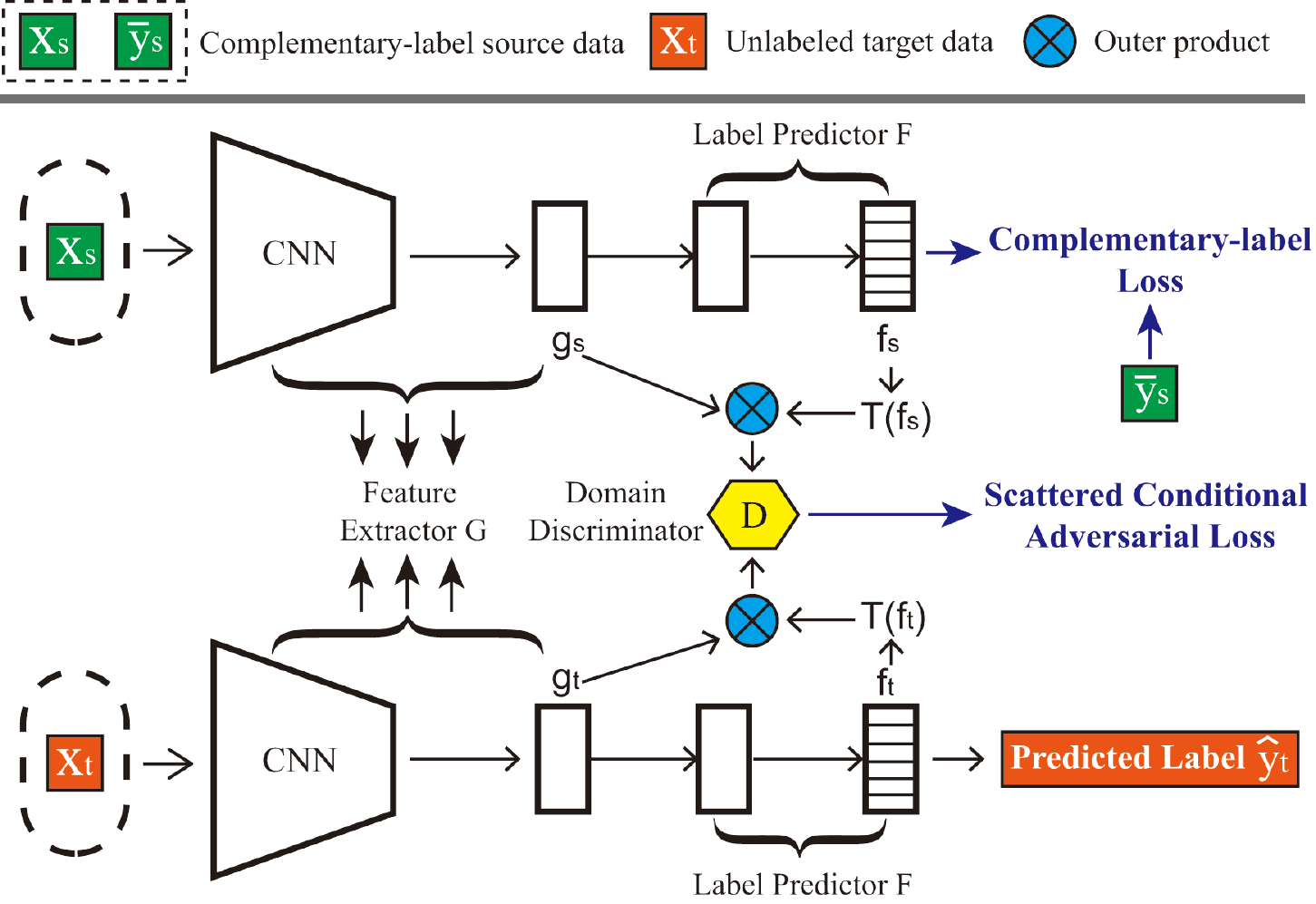}}
		\end{center}
		\caption{{Overview of the proposed \emph{\underline{c}omplementary \underline{l}abel advers\underline{ari}al \underline{net}work} (CLARINET). It consists of feature extractor $G$, label predictor $F$ and conditional domain discriminator $D$. $g_s$ and $g_t$ are the outputs of $G$, representing the extracted features of source and target data. $f_s$ and $f_t$ represent classifier predictions. $T$ is a mapping function which we propose to scatter the classifier predictions. In Algorithm~\ref{alg:algorithm}}, we show how to use two losses mentioned in this figure to train CLARINET.}
		\label{fig: CLARINET}
\end{figure}

While determining the correct label from many candidates is laborious, choosing one of the incorrect labels (i.e., complementary labels) would be much easier and thus less costly, especially when we have many candidates \cite{ishida2017learning}. For example, suppose that we need to annotate labels of a bunch of animal images from $1,000$ candidates. One strategy is to ask crowd-workers to choose the true labels from $1,000$ candidates, while the other is to judge the correctness of a label randomly given by the system from the candidates. Apparently, the cost of the second strategy is much lower than that of the first one \cite{pmlr-v97-ishida19a}. 

This brings us a novel problem setting, {\em budget-friendly UDA} (abbreviated as BFUDA), which aims to transfer knowledge from complementary-label source data to unlabeled target data (Figure~\ref{fig: our_solution}). We describe this problem setting using the word {\em budget-friendly} since, compared to ordinary UDA, we can greatly save the labeling cost by annotating complementary labels in the source domain rather than annotating true labels  \cite{ishida2017learning,yu2018learning}. Please note that, existing UDA methods cannot handle BFUDA, as they require fully-true-label source data \cite{Liu2017,saito2017asymmetric} or at least $20\%$ true-label source data (demonstrated in  \cite{liu2019butterfly,TCL_long}).

A straightforward but weak solution to BFUDA is a two-step approach, which sequentially combines \emph{complementary-label learning} (CLL) methods and existing UDA methods (green dash line in Figure~\ref{fig: our_solution})\footnote{We implement this two-step approach and take it as a baseline.}. CLL methods are used to assign pseudo labels for complementary-label source data. Then, we can train a target-domain classifier with pseudo-label source data and unlabeled target data using existing UDA methods. Nevertheless, pseudo-label source data contain noise, which may cause poor domain-adaptation performance of such two-step approach \cite{liu2019butterfly}.

Therefore, we propose a powerful one-step solution to BFUDA,  \emph{\underline{c}omplementary \underline{l}abel advers\underline{ari}al \underline{net}work} (CLARINET). It maintains two deep networks trained by adversarial way simultaneously, where one can accurately classify complementary-label source data, and the other can discriminate source and target domains. Since Long et. al. \shortcite{long2018conditional} and Song et. al. \shortcite{song2009hilbert} have shown that multimodal structures of distributions can only be captured sufficiently by the cross-covariance dependency between the features and classes (i.e., true labels), we set the input of domain discriminator $D$ as the outer product of feature representation (e.g., $\bm{g}_s$ in Figure~\ref{fig: CLARINET}) and mapped classifier prediction (e.g., $\bm{T(f_s)}$ in Figure~\ref{fig: CLARINET}).

Due to the nature of complementary-label classification, predicted probability of each class (i.e., each element of $\bm{f_s}$, Figure~\ref{fig: CLARINET}) is relatively close.
According to \cite{song2009hilbert}, this kind of predicted probabilities could not provide sufficient information to capture the multimodal structure of distributions. To fix it, we propose a mapping function $T$ to make the predicted probabilities more scattered (i.e., $\bm{T(f_s)}$, Figure~\ref{fig: CLARINET}) than previous ones (i.e., $\bm{f_s}$, Figure~\ref{fig: CLARINET}). By doing so, the mapped classifier predictions can better indicate their choice. In this way, we can take full advantage of classifier predictions and effectively align distributions of two domains. Our ablation study (see Table~\ref{tab:ablation}) verifies that $T$ indeed helps improve the target-domain accuracy.

We conduct experiments on $6$ BFUDA tasks and compare CLARINET with a series of competent baselines. Empirical results demonstrated that CLARINET can effectively transfer knowledge from complementary-label source data to unlabeled target data and is superior to all baselines.

\section{Budget-friendly Unsupervised Domain Adaptation}

In this section, we propose a novel problem setting, called \emph{budget-friendly unsupervised domain adaptation} (BFUDA), and prove a learning bound of this new problem. Then, we show how BFUDA brings benefits to domain adaptation field.



\subsection{Problem Setting}

Let $\mathcal{X}\subset \mathbb{R}^d$ be a feature (input) space and $\mathcal{Y}:=\{1,...,K\}$ be a label (output) space. A \textit{domain} is defined as follows.
\begin{definition}[Domains for BFUDA]\label{d3}Given random variables $X_s, X_t \in \mathcal{X}$, $Y_s, \overline{Y}_s, {Y}_t \in \mathcal{Y}$, the {source} and {target domains} are joint distributions $P(X_s, \overline{Y}_s)$ and $P(X_t, {Y}_t)$, where the joint distributions $P(X_s, Y_s)\neq P(X_t, {Y}_t)$ and $P(\overline{Y}_s=c|Y_s=c)=0$ for all $c\in \mathcal{Y}$.
\end{definition}
Then, we propose BFUDA problem as follows.
\begin{definition*}[BFUDA]
Given independent and identically distributed (\textit{i.i.d.})~labeled samples $\overline{D}_s=\{(\mathbf{x}_s^i,\overline{y}_s^i)\}^{\overline{n}_s}_{i=1}$ drawn from the source domain $P(X_s, \overline{Y}_s)$ and \textit{i.i.d.}~unlabeled samples $D_t=\{\mathbf{x}_t^i\}^{n_t}_{i=1}$ drawn from the target marginal distribution $P(X_t)$, {the aim} of BFUDA is to train a classifier  $F:\mathcal{X}\rightarrow \mathcal{Y}$ with $\overline{D}_s$ and $D_t$ such that
$F$ can accurately classify target data drawn from $P(X_t)$.
\end{definition*}
It is clear that it is impossible to design a
suitable learning procedure without any assumptions on $P(X_s, \overline{Y}_s)$. In this paper, we use the assumption for unbiased
complementary learning proposed by \cite{pmlr-v97-ishida19a,ishida2017learning}:


\begin{equation}
\label{assumption}
    P(\overline{Y}_s=k|X_s)=\frac{1}{K-1}\sum_{c=1, c\neq k}^K P({Y}_s=c|X_s),
\end{equation}
for all $k, c \in \mathcal{Y}=\{1,...,K\}$ and $c\neq k$.

\subsection{Learning Bound of BFUDA}
\label{Theory}

This subsection presents a learning bound of BFUDA. Practitioner may safely skip it.
The $\ell:\mathbb{R}^K\times \mathcal{Y} \rightarrow \mathbb{R}_+$ is a loss function. The \emph{decision function} is a vector-valued function ${\bm h}: \mathcal{X}\rightarrow \mathbb{R}^K$ and ${\bm h}_k$ is the $k$-th element of ${\bm h}$. The complementary risk for ${\bm h}$ with respect to $\ell$
over $P({{X}_s,\overline{Y}_s})$ is
\begin{equation*}
    {L}_{\overline{s}}({\bm h})=\mathbb{E}\ell({\bm h}(X_s),\overline{Y}_s).
\end{equation*}
The risks for the decision function ${\bm h}$ with respect to loss $\ell$
over implicit distribution $P({{X}_s,{Y}_s}), P({{X}_t,{Y}_t})$ are:
\begin{equation*}
\begin{split}
& L_s({\bm h})=\mathbb{E}\ell({\bm h}(X_s),{Y}_s),
L_t({\bm h})=\mathbb{E}\ell({\bm h}(X_t),{Y}_t).
\end{split}
\end{equation*}
Then, we introduce our main theorem as follows.
\begin{theorem}\label{Mainthe}
Given a loss function $\ell$ and  a hypothesis $\mathcal{H} \subset \{{\bm h}:\mathcal{X}\rightarrow \mathbb{R}^K\}$, then under unbiased assumption, for any ${\bm h}\in \mathcal{H}$,  we have 
\begin{equation*}
\begin{split}
L_t({\bm h})&\leq  \overline{L}_s({\bm h}) +\frac{1}{2}d_{\mathcal{H}}^{\ell}(P_{{X}_s},P_{{X}_t})+\Lambda,
 \end{split}
\end{equation*}
where $
\overline{L}_s({\bm h}):=\sum_{k=1}^K\int_{\mathcal{X}} \ell({\bm h}(\mathbf{x}),k) {\rm d}P_{{X}_s}-(K-1) {L}_{\overline{s}}({\bm h})$, $P_{X_s}, P_{X_t}$ are source and target marginal distributions, $\Lambda={\min}_{{\bm h}\in \mathcal{H}}  ~R_s({\bm h})+R_t({\bm h})$, and $d_{\mathcal{H}}^{\ell}(P_{{X}_s},P_{{X}_t})$ is the distribution discrepancy defined in \cite{ben2010theory}.

\label{theorem1}
\end{theorem}
\begin{proof}
We firstly investigate the connection between $L_s({\bm h})$ and ${L}_{\overline{s}}({\bm h})$ under unbiased assumption in Eq.~\eqref{assumption}.
Given $K\times K$ matrix $Q$ whose diagonal elements are $0$ and other elements are $1/K$, we represent the unbiased assumption by $\overline{\bm {\bm \eta}}=Q{{\bm \eta}}$, 
where $\overline{{\bm \eta}}=[P(\overline{Y}_s=1|X_s),...,P(\overline{Y}_s=K|X_s)]^T$ and ${{\bm \eta}}=[P({Y}_s=1|X_s),...,P({Y}_s=K|X_s)]^T$.
Note that $Q$ has inverse matrix $Q^{-1}$ whose diagonal elements are $-(K-2)$ and other elements are $1$. Thus, we have that
\begin{equation}\label{mainequation}
     Q^{-1}\overline{{\bm \eta}}={{\bm \eta}}.
\end{equation}
According to Eq.~\eqref{mainequation}, we have
    $P(Y_s=k|X_s) = 1-(K-1)P(\overline{Y}_s=k|X_s)$,
which implies that
\begin{equation}\label{main}
\begin{split}
    L_s({\bm h})&=\sum_{k=1}^K\int_{\mathcal{X}} \ell({\bm h}(\mathbf{x}),k) {\rm d}P_{{X}_s}-(K-1) {L}_{\overline{s}}({\bm h}).
\end{split}
\end{equation}
Hence, $L_s({\bm h})=\overline{L}_s({\bm h})$.
Combining Eq.~\eqref{main} with the domain adaptation bound presented in \cite{ben2010theory}
\begin{equation*}
\begin{split}
L_t({\bm h})&\leq   L_s({\bm h}) +\frac{1}{2}d_{\mathcal{H}}^{\ell}(P_{{X}_s},P_{{X}_t})+\Lambda,
 \end{split}
\end{equation*}
we prove this theorem.
\end{proof}
In this bound, the first term, i.e., Eq. \eqref{main}, is the source classification error based on complementary-label source data. 
The second term is the distribution discrepancy distance between two domains and $\Lambda$ is the difference in labeling functions across the two domains. The empirical form of Eq. \eqref{main} is known as complementary-label loss (see Eq.~\eqref{E2}).

\subsection{How does BFUDA Bring Benefits to DA Field?}
Collecting true-label data is always expensive in the real world. Thus, learning from less expensive data \cite{kumar2017semi,sakai} has been extensively studied in machine learning field, including label-noise learning \cite{han2018co,han2018masking}, complementary-label learning \cite{pmlr-v97-ishida19a,yu2018learning,ishida2017learning} and so on. Among all these research directions, obtaining complementary labels is a cost-effective option. As described in the previous works mentioned above, compared with choosing the true class out of many candidate classes precisely, collecting complementary labels is obviously much easier and less costly. In addition, a classifier trained with complementary-label data is equivalent to a classifier trained with true-label data as shown in \cite{pmlr-v97-ishida19a}.

At present, the success of DA still highly relies on the scale of true-label source data, which is a critical bottleneck. With limited budget, it is unrealistic to obtain enough true-label source data and thus cannot achieve a good distribution adaptation result. For the same budget, we can get multiple times more complementary-label data than the true-label data. In addition, the adaptation scenario is limited to some commonly used datasets, as they have sufficient true labels to support distributional adaptation. Thus if we can reduce the labeling cost in the source domain, for example, by using complementary-label data to replace true-label data (i.e. BFUDA), we can promote DA to more fields. Due to existing UDA methods require at least $20\%$ true-label source data \cite{TCL_long}, they cannot handle BFUDA problem. To address BFUDA problem directly, we propose a powerful solution, CLARINET, as follows.

\section{CLARINET: One-step BFUDA Approach}

The proposed CLARINET (Figure \ref{fig: CLARINET}) mainly consists of feature extractor $G$, label predictor $F$ and domain discriminator $D$. Furthermore, we add a mapping function $T$ between the label predictor $F$ and domain discriminator $D$ to take full advantage of classifier predictions. In this section, we first introduce two losses used to train CLARINET, and then the whole training procedures of CLARINET is presented.

\subsection{Loss Function in CLARINET}
\label{sec:Loss}
In this subsection, we show how to compute the two losses mentioned above in CLARINET after obtaining mini-batch $\overline{d}_s$ from $\overline{D}_s$ and  $d_t$ from $D_t$.


\paragraph{Complementary-label Loss.} 
We first divided $\overline{d}_s$ into $K$ disjoint subsets according to the complementary labels in $\overline{d}_s$,
\begin{equation}
    \overline{d}_s=\cup_{k=1}^K \overline{d}_{s,k},~\overline{d}_{s,k}=\{(\mathbf{x}_k^i,k)\}_{i=1}^{\overline{n}_{s,k}},
\label{E1}
\end{equation}
where $\overline{d}_{s,k} \cap \overline{d}_{s,k'}=\varnothing$ if $k\neq{k'}$ and $\overline{n}_{s,k}=|\overline{d}_{s,k}|$. Then, following Eq. \eqref{main}, the complementary-label loss on $\overline{d}_{s,k}$ is
\begin{equation}
\begin{split}
    {\overline{L}}_{s}(G,F,\overline{d}_{s,k})=-&(K-1)\frac{{\overline\pi}_k}{\overline{n}_{s,k}}\sum_{i=1}^{\overline{n}_{s,k}}\ell(F\circ{G}(\mathbf{x}_k^i),k) \\
    +&\sum_{j=1}^{K}\frac{{\overline\pi}_j}{\overline{n}_{s,j}}\sum_{l=1}^{\overline{n}_{s,j}}\ell(F\circ{G}(\mathbf{x}_j^l),k),
\end{split}
\label{E2}
\end{equation}
where $\ell$ can be any loss and we use the cross-entropy loss, ${\overline\pi}_k$ is the proportion of the samples complementary-labeled $k$. The total complementary-label loss on $\overline{d}_s$ is as follows.
\begin{equation}
    {\overline{L}}_s(G,F,\overline{d}_s)=\sum_{k=1}^K{\overline{L}}_s(G,F,\overline{d}_{s,k}).
\label{E3}
\end{equation}


As shown in Section \ref{Theory}, the complementary-label loss (i.e., Eq.~\eqref{E3}) is an unbiased estimator of the true-label-data risk.  Namely, the minimizer of complementary-label loss agrees with the minimizer of the true-label-data risk with no constraints on the loss $\ell$ and model $F\circ{G}$ \cite{pmlr-v97-ishida19a}.

\begin{remark}
Due to the negative part in ${\overline{L}}_s(G,F,\overline{d}_s)$,  minimizing it will cause over-fitting \cite{kiryo2017positive}. To overcome this problem, we use a correctional way \cite{pmlr-v97-ishida19a} to minimize ${\overline{L}}_s(G,F,\overline{d}_s)$ (lines $7$-$13$ in Algorithm \ref{alg:algorithm}).
\end{remark}


\paragraph{Scattered Conditional Adversarial Loss.} 

According to \cite{song2009hilbert}, it is significant to capture multimodal structures of distributions using cross-covariance dependency between the features and classes (i.e., true labels). Since there are no true-label target data in UDA, CDAN adopts outer product of feature representations and classifier predictions (i.e., outputs of the softmax layer) as new features of two domains \cite{long2018conditional}. The newly constructed features have shown great ability to discriminate source and target domains, since classifier predictions of true-label source data are dispersed, expressing the predicted goal clearly.

However, in the complementary-label classification mode, we observe that the predicted probability of each class (i.e., each element of $\bm{f_s}$ in Figure~\ref{fig: CLARINET})
is relatively close. Namely, it is hard to find significant predictive preference from the classifier predictions. According to \cite{song2009hilbert}, this kind of predictions cannot provide sufficient information to capture the multimodal structure of distributions. To fix it, we propose a mapping function $T$ to scatter the classifier predictions ${\bm f}=[f_1,...,f_K]^T$ ($\bm f$ could be $\bm{f_s}$ or $\bm{f_t}$ in Figure~\ref{fig: CLARINET}),
\begin{equation}
    T({\bm f})=\left [\frac{{f}_1^{\frac{1}{l}}}{\sum_{j=1}^{K}{f}_j^{\frac{1}{l}}},..,\frac{{f}_k^{\frac{1}{l}}}{\sum_{j=1}^{K}{f}_j^{\frac{1}{l}}},...,\frac{{f}_K^{\frac{1}{l}}}{\sum_{j=1}^{K}{f}_j^{\frac{1}{l}}} \right ]^T.
\label{eq:T1}
\end{equation}
It is a common approach of adjusting the “temperature” of categorical distribution. As $l \rightarrow 0$, the output of $T({\bm f})$ will approach a Dirac (“one-hot”) distribution.

Then to prioritize the discriminator on those easy-to-transfer examples, following \cite{long2018conditional}, we measure the uncertainty of the prediction for sample $\mathbf{x}$ by
\begin{equation}
    H(G,F,\mathbf{x})=-\sum_{k=1}^K{T(f_k(\mathbf{x}))}{\log{T(f_k(\mathbf{x}))}}.
\label{EH}
\end{equation}
Thus the scattered conditional adversarial loss is as follows,
\begin{equation}
\begin{split}
    \displaystyle
    L_{adv}(G,F,D,\overline{d}_s,d_t)=&\frac{\sum_{\mathbf{x}\in \overline{d}_{s}[X]}\omega_{\overline{s}}(\mathbf{x})\log(D({\bm g(\mathbf{x})}))}{\sum_{\mathbf{x}\in \overline{d}_{s}[X]}\omega_{\overline{s}}(\mathbf{x})}
    \\+&\frac{\sum_{\mathbf{x}\in d_t}\omega_t(\mathbf{x}){\log(1-D({\bm g(\mathbf{x})}))}}{\sum_{\mathbf{x}\in d_t}\omega_t(\mathbf{x})},
\end{split}
\label{adv1}
\end{equation}
where $\omega_{\overline{s}}(\mathbf{x})$ and $\omega_{t}(\mathbf{x})$ are $1+e^{-H(G,F,\mathbf{x})}$, ${\bm g}(\mathbf{x})$ is $G(\mathbf{x})\otimes{T(F\circ{G}(\mathbf{x})})$ and $\overline{d}_{s}[X]$ is the feature part of $\overline{d}_s$.

\begin{figure*}[t]
\centering
\subfigure[USPS $\rightarrow$ MNIST.]{
\includegraphics[width=0.331\textwidth]{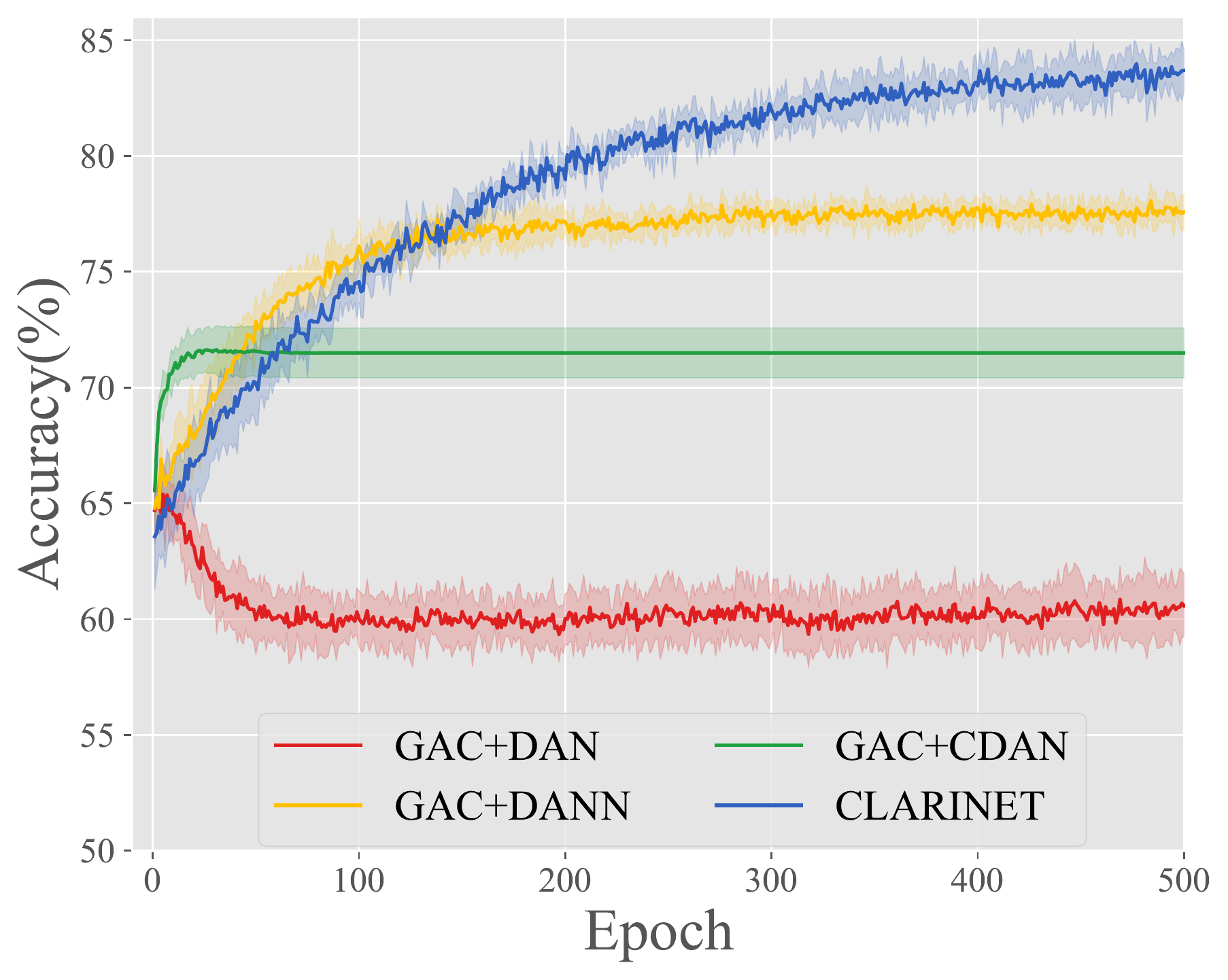}
}
\subfigure[MNIST $\rightarrow$ USPS.]{
\includegraphics[width=0.314\textwidth]{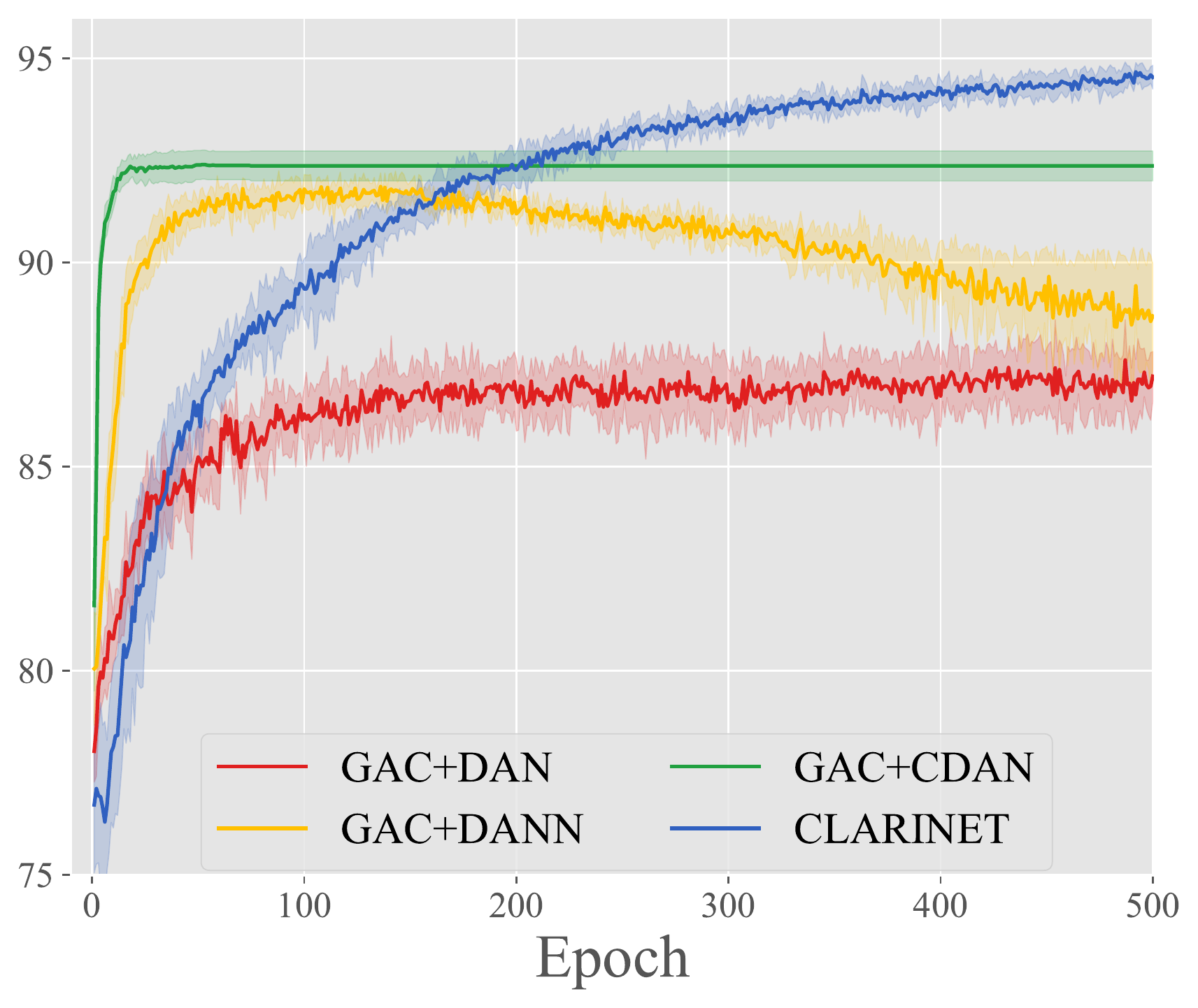}
}
\subfigure[SVHN $\rightarrow$ MNIST.]{
\includegraphics[width=0.314\textwidth]{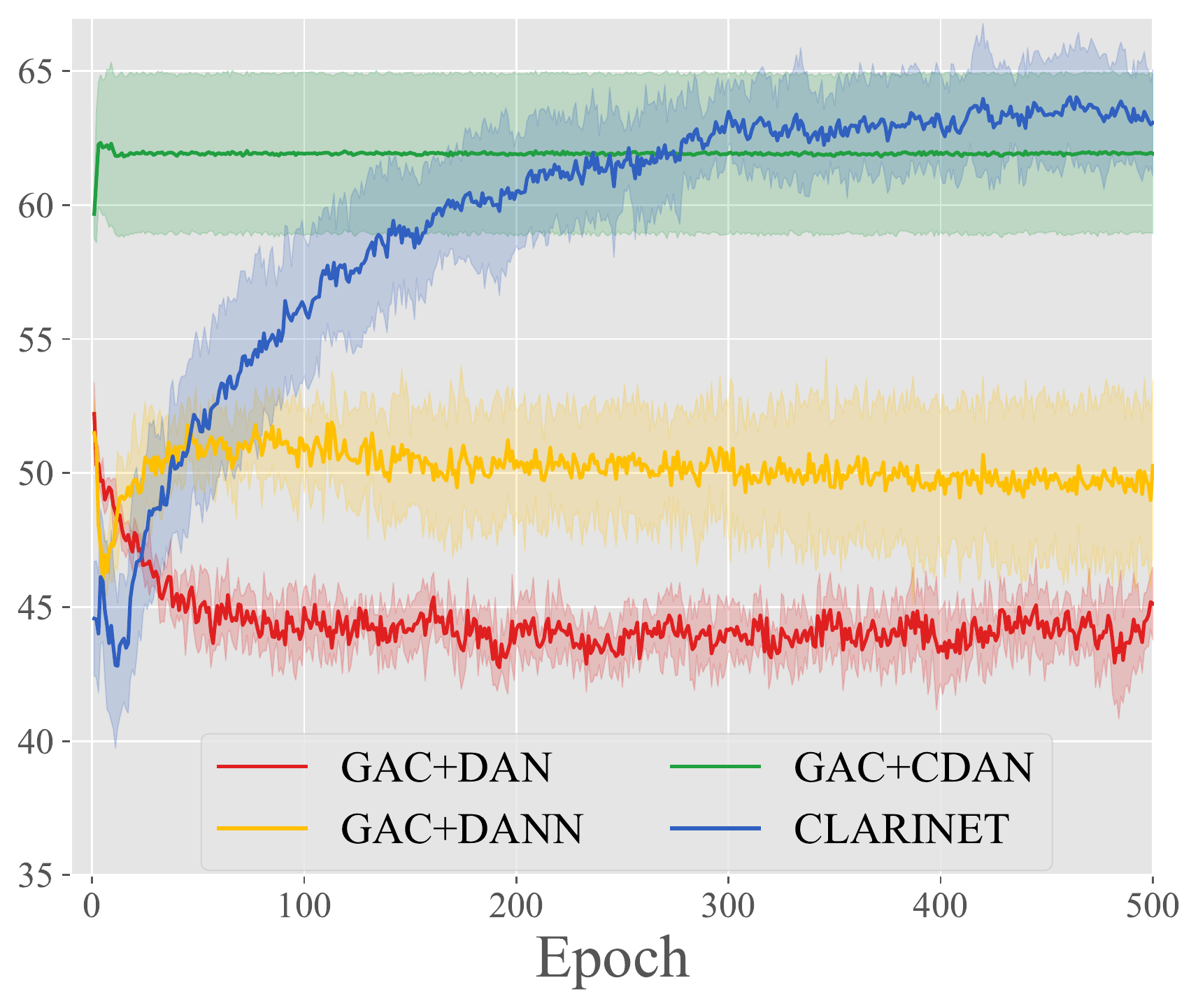}
}
\subfigure[MNIST $\rightarrow$ MNIST-M.]{
\includegraphics[width=0.331\textwidth]{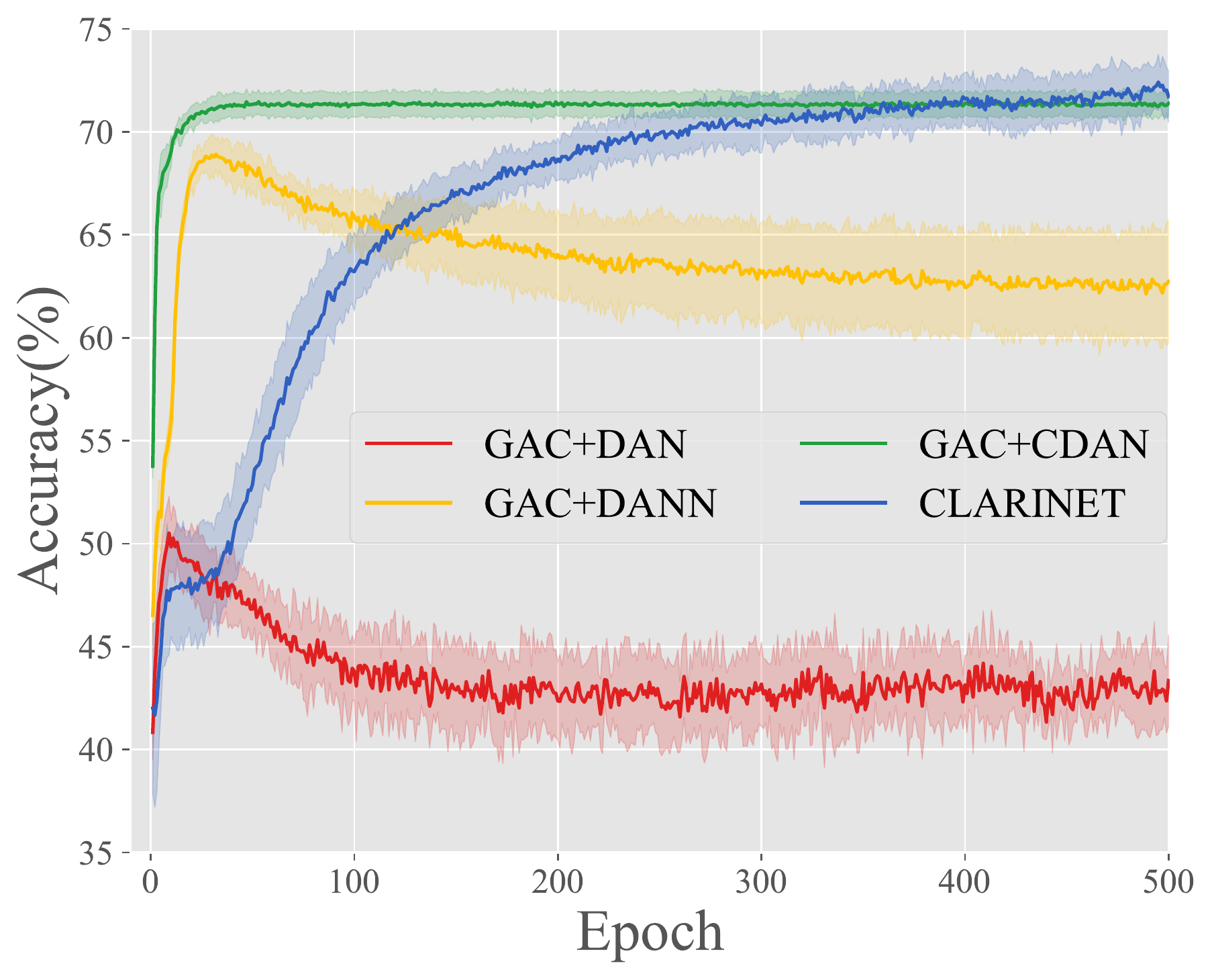}
}
\subfigure[SYND $\rightarrow$ MNIST.]{
\includegraphics[width=0.314\textwidth]{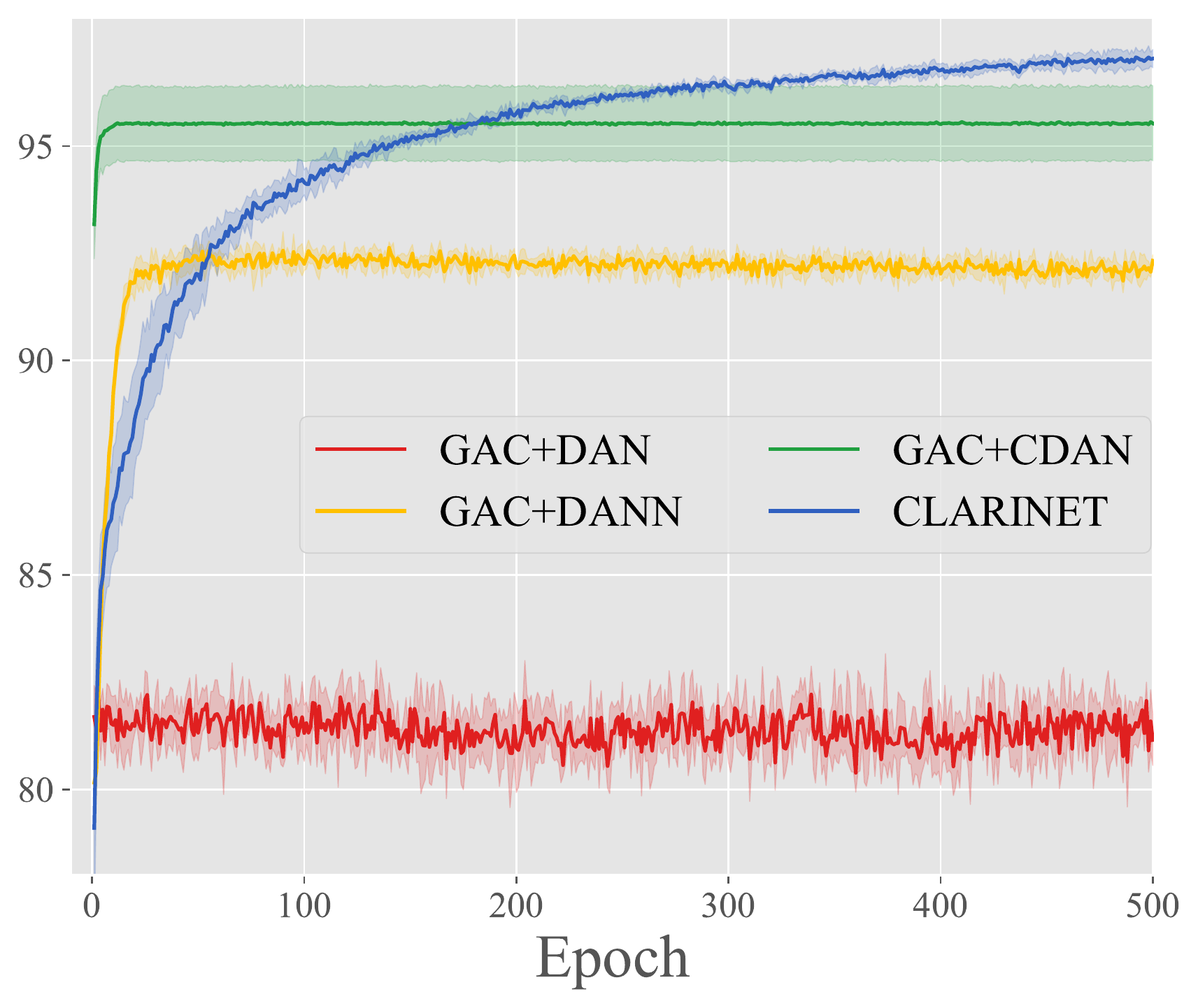}
}
\subfigure[SYND $\rightarrow$ SVHN.]{
\includegraphics[width=0.314\textwidth]{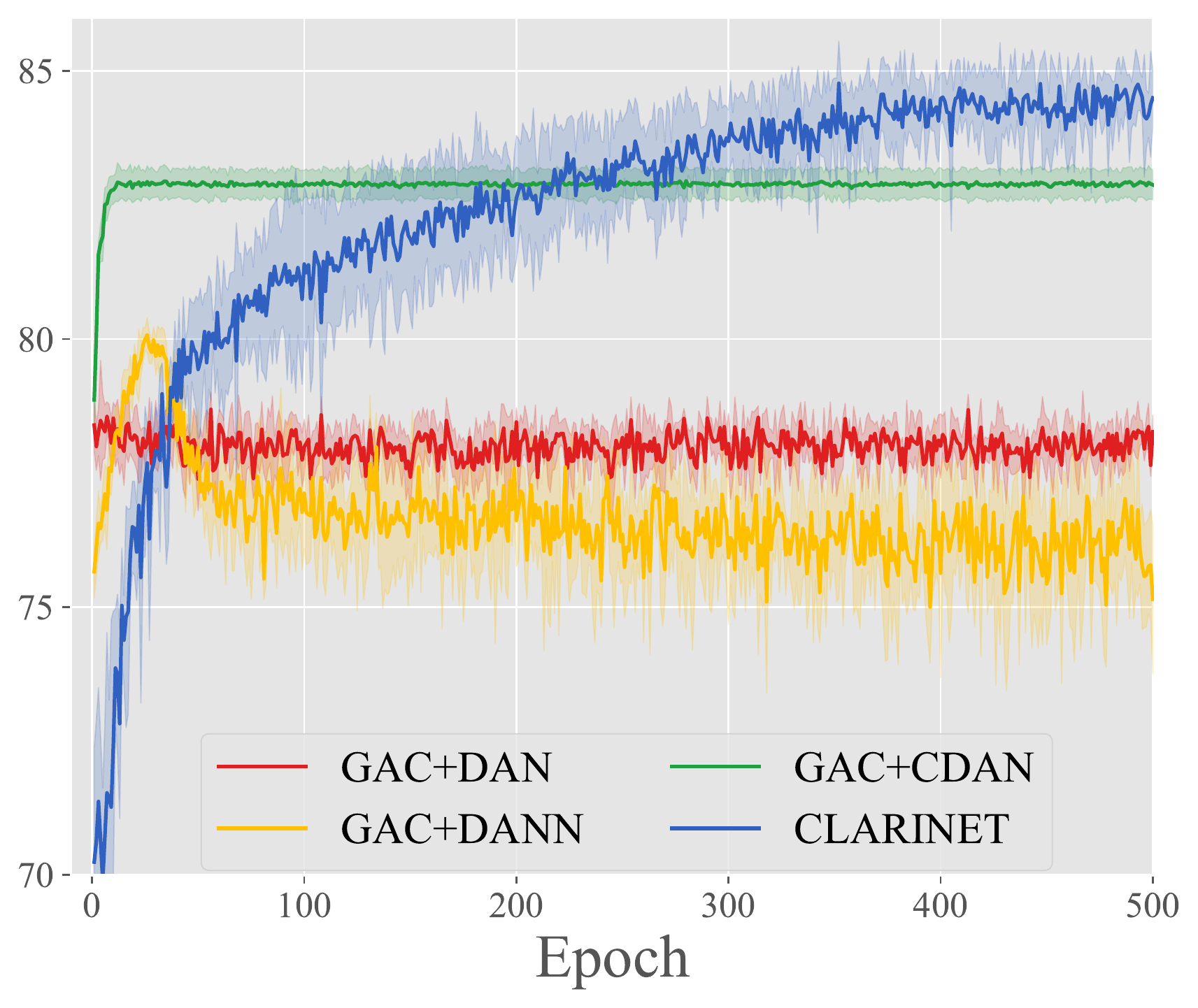}
}
\caption{Test Accuracy vs. Epochs on $6$ BFUDA Task. In (a)-(f), we compare target-domain accuracy of one-step approach, i.e., CLARINET (\emph{ours}), with that of two-step approach (\emph{ours}). It is clear to see that the accuracy of CLARINET gradually and steadily increases and eventually converges, achieving the best accuracy on each task.} 
\label{fig:ccom}
\end{figure*}

\begin{algorithm}[tb]
\small
\caption{CLARINET: One-step BFUDA Approach}
\label{alg:algorithm}
\textbf{Input}: $\overline{D}_s=\{(\mathbf{x}_s^i,\overline{y}_s^i)\}^{\overline{n}_s}_{i=1}$, $D_t=\{\mathbf{x}_t^i\}^{n_t}_{i=1}$.\\
\textbf{Parameters}: learning rate $\gamma_1$ and $\gamma_2$, epoch $T_{max}$, start epoch $T_s$, iteration $N_{max}$, class number $K$, tradeoff $\lambda$, network parameter $\theta_{F\circ{G}}$ and $\theta_D$.\\
\textbf{Output}: the neural network $F\circ{G}$, namely the target domain classifier for $D_t$.
\begin{algorithmic}[1] 
\STATE \textbf{Initialize} $\theta_{F\circ{G}}$ and $\theta_D$;
\FOR{$t=1,2……T_{max}$}
\STATE \textbf{Shuffle} the training set $\overline{D}_s$, ${D}_t$;
\FOR{$N=1,2……N_{max}$}
\STATE \textbf{Fetch} mini-batch $\overline{d}_s$, $d_t$ from $\overline{D}_s$, ${D}_t$;
\STATE \textbf{Divide} $\overline{d}_s$ into $\{\overline{d}_{s,k}\}_{k=1}^K$;
\STATE \textbf{Calculate} $\{\overline{L}_s(G,F,\overline{d}_{s,k})\}_{k=1}^K$ using Eq.~\eqref{E2}, and $\overline{L}_s(G,F,\overline{d}_{s})$ using Eq.~\eqref{E3};
\IF{$\min_k\{\overline{L}_s(G,F,\overline{d}_{s,k})\}_{k=1}^{K}\geq 0$}
\STATE \textbf{Update} $\theta_{F\circ{G}}=\theta_{F\circ{G}}-\gamma_1\triangledown \overline{L}_s(G,F,\overline{d}_{s})$;
\ELSE
\STATE \textbf{Calculate} $\overline{L}_{neg}=\sum_{k=1}^K\min\{0,\overline{L}_s(G,F,\overline{d}_{s,k})\}$;
\STATE \textbf{Update} $\theta_{F\circ{G}}=\theta_{F\circ{G}}+\gamma_1\triangledown \overline{L}_{neg}$;
\ENDIF
\IF{$t > T_s$}
\STATE \textbf{Calculate} $L_{adv}(G,F,D,\overline{d}_s,d_t)$ using Eq.~\eqref{adv1};
\STATE \textbf{Update} $\theta_D=\theta_D-\gamma_2\triangledown{L_{adv}(G,F,D,\overline{d}_s,d_t)}$;
\STATE \textbf{Update} $\theta_{F\circ{G}}=\theta_{F\circ{G}}+\gamma_2\lambda\triangledown{L_{adv}(G,F,D,\overline{d}_s,d_t)}$;
\ENDIF
\ENDFOR
\ENDFOR
\end{algorithmic}
\end{algorithm}

\subsection{Training Procedure of CLARINET}

Based on the two losses proposed in Section~\ref{sec:Loss}, in CLARINET, we try to solve the following optimization problem,
\begin{equation}
\label{eq:minmax}
\begin{split}
    \min_{G,F}&~\overline{L}_s(G,F,\overline{D}_s)-\lambda L_{adv}(G,F,D,\overline{D}_s,D_t),\\
    \min_D&~ L_{adv}(G,F,D,\overline{D}_s,D_t),
\end{split}
\end{equation}
where $D$ tries to distinguish the samples from different domains by minimizing $L_{adv}$, while $F\circ{G}$ wants to maximize the $L_{adv}$ to make domains indistinguishable.   
To solve the minimax optimization problem in Eq.~\eqref{eq:minmax}, we add a gradient reversal layer \cite{ganin2016domain} between the domain discriminator and the classifier, which multiplies the gradient by a negative constant (-$\lambda$) during the back-propagation. $\lambda$ is a hyper-parameter between the two losses to tradeoff source risk and domain discrepancy. 

The training procedures of CLARINET are shown in Algorithm~\ref{alg:algorithm}. First, we initialize the whole network (line $1$) and shuffle the training set (line $3$). During each epoch, after minbatch $\overline{d}_s$ and $d_t$ are fetched (line $5$), we divide the source mini-batch $\overline{d}_s$ into $\{\overline{d}_{s,k}\}_{k=1}^K$ using Eq.~\eqref{E1} (line $6$). Then, $\{\overline{d}_{s,k}\}_{k=1}^K$ are used to calculate the complementary-label loss for each class (i.e., $\{\overline{L}_s(G,F,\overline{d}_{s,k})\}_{k=1}^{K}$) and the whole complementary-label loss $\overline{L}_s(G,F,\overline{d}_s)$ (line $7$).

If $\min_k\{\overline{L}_s(G,F,\overline{d}_{s,k})\}_{k=1}^{K}\geq 0$, we calculate the gradient $\triangledown \overline{L}_s(G,F,\overline{d}_s)$ and update parameters of $G$ and $F$ using gradient descent (lines $8$-$9$). 
Otherwise, we sum negative elements in $\{\overline{L}_s(G,F,\overline{d}_{s,k})\}_{k=1}^{K}$ as $\overline{L}_{neg}$ (line $11$) and calculate the gradient with $\triangledown\overline{L}_{neg}$ (line $12$). 
Then, we update parameters of $G$ and $F$ using gradient ascent (line $12$), which is suggested by \cite{pmlr-v97-ishida19a}. 
When the number of epochs (i.e., $t$) is over $T_s$, we start to update parameters of $D$ (line $14$). We calculate the scattered conditional adversarial loss $L_{adv}$ (line $15$). Then, $L_{adv}$ is minimized over $D$ (line $16$), but maximized over $F\circ{G}$ (line $17$) for adversarial training. 

\section{Experiments}


Based on five commonly used datasets: {\emph{MNIST}} (\emph{M}), {\emph{USPS}} (\emph{U}), {\emph{SVHN}} (\emph{S}), {\emph{MNIST-M}} (\emph{m}) and {\emph{SYN-DIGITS}} (\emph{Y}), we verify efficacy of CLARINET on $6$ {BFUDA tasks}: $\emph{M}$ $\rightarrow$ $\emph{U}$, $\emph{U}$ $\rightarrow$ $\emph{M}$, $\emph{S}$ $\rightarrow$ $\emph{M}$,  $\emph{M}$ $\rightarrow$ $\emph{m}$, $\emph{Y}$$\rightarrow$ $\emph{M}$ and $\emph{Y}$$\rightarrow$ $\emph{S}$. Note that, we generate complementary-label data according to \cite{pmlr-v97-ishida19a}.

\subsection{Baselines}

We compare CLARINET with the following baselines: {\em gradient ascent complementary label learning} (GAC) \cite{pmlr-v97-ishida19a}, namely non-transfer method, and several two-step methods, which sequentially combine GAC with UDA methods (including DAN \cite{long2015learning}, DANN \cite{ganin2016domain} and CDAN \cite{long2018conditional}). In two-step approach, GAC method is first used to assign pseudo labels for complementary-label source data. Then, we train the classifier with pseudo-label source data and unlabeled target data using UDA methods. Thus, we have four possible baselines: GAC, GAC+DAN, GAC+DANN and GAC+CDAN. For two-step methods, they share the same pseudo-label source data on each task. Note that, in this paper, we use the entropy conditioning variant of CDAN.  


\begin{table*}[t]
  \centering
  \small
    \setlength{\tabcolsep}{6mm}{
    \begin{tabular}{lccccc}
    \toprule
    \multirow{2}[4]{*}{Tasks} & \multirow{2}[4]{*}{GAC} & \multicolumn{3}{c}{Two-step approaches (\emph{ours})} & \multirow{1}[4]{*}{CLARINET} \\
\cmidrule{3-5}          &       & GAC+DAN & GAC+DANN & GAC+CDAN\_E & (\emph{ours}) \\
    \midrule
    $U \rightarrow M$   & 51.860  & 60.692$\pm$1.300  & 77.580$\pm$0.770  & 71.498$\pm$1.077  & \textbf{83.692$\pm$0.928 } \\
    $M \rightarrow U$   & 77.796  & 87.215$\pm$0.603  & 88.688$\pm$1.280  & 92.366$\pm$0.365  & \textbf{94.538$\pm$0.292 } \\
    $S \rightarrow M$   & 39.260  & 45.132$\pm$1.363  & 50.882$\pm$2.440  & 61.922$\pm$2.983  & \textbf{63.070$\pm$1.990 }  \\
    $M \rightarrow m$   & 45.045  & 43.346$\pm$2.224  & 62.273$\pm$2.261  & 71.379$\pm$0.620  & \textbf{71.717$\pm$1.262 } \\
    $Y \rightarrow M$   & 77.070  & 81.150$\pm$0.591  & 92.328$\pm$0.138  & 95.532$\pm$0.873  & \textbf{97.040$\pm$0.212 } \\
    $Y \rightarrow S$   & 72.480  & 78.270$\pm$0.311  & 75.147$\pm$1.401  & 82.878$\pm$0.278  & \textbf{84.499$\pm$0.537 } \\
    \midrule
    Average  & 60.585  & 65.968   & 74.483  & 79.263  & \textbf{82.426 } \\
    \bottomrule
    \end{tabular}}%
  \caption{Results on $6$ BFUDA Tasks. Bold value represents the highest accuracy (\%) on each row. Please note, the two-step methods and CLARINET are all first proposed in our paper. }
  \label{tab:ccom}%
\end{table*}%

\begin{table*}[htbp]
  \centering
  \small
  \resizebox{\textwidth}{!}
  {
    \begin{tabular}{lccccccc}
    \toprule
    Methods & $U \rightarrow M$   & $M \rightarrow U$   & $S \rightarrow M$   & $M \rightarrow m$   & $Y \rightarrow M$   & $Y \rightarrow S$   & Average \\
    \midrule
    C w/ $L_{CE}$   & 0.445$\pm$0.722  & 0.055$\pm$0.129  & 3.708$\pm$0.688  & 7.088$\pm$0.424  & 1.832$\pm$0.102  & 1.298$\pm$0.070  & 2.404  \\
    C w/o $T$ & 83.192$\pm$1.796  & 93.419$\pm$0.588  & 52.438$\pm$1.927  & \textbf{ 72.128$\pm$1.569 } & 95.442$\pm$1.004  & 83.055$\pm$0.652  & 79.946  \\
    CLARINET & \textbf{ 83.692$\pm$0.928 } & \textbf{ 94.538$\pm$0.292 } & \textbf{ 63.070$\pm$1.990 } & 71.717$\pm$1.262  & \textbf{ 97.040$\pm$0.212 } & \textbf{ 84.499$\pm$0.537 } & \textbf{ 82.426 } \\
    \bottomrule
    \end{tabular}}%
    \caption{Ablation Study. Bold value represents the highest accuracy (\%) on each column. We prove UDA methods cannot handle BFUDA tasks directly and the mapping function $T$ can help improve the adaptation performance under BFUDA.}
  \label{tab:ablation}%
\end{table*}%

\subsection{Experimental Setup}
\label{Asec:Exp_set}

We design our feature extractor $G$, label predictor $F$ and domain discriminator $D$ according to the architecture from previous works. More precisely, we pick the structures of feature extractor $G$ from \cite{ganin2016domain,long2018conditional}. The label predictor $F$ and domain discriminator $D$ all share the same structure in all tasks, following CDAN \cite{long2018conditional}. We follow the standard protocols for unsupervised domain adaptation and compare the average classification accuracy based on $5$ random experiments. For each experiment, we take the result of the last epoch. The batch size is set to $128$ and the number of epochs is set to $500$. SGD optimizer (momentum = $0.9$, weight\_decay = $5\times{10^{-5}}$) is with an initial learning rate of  $0.005$ in adversarial network and $5\times{10^{-5}}$ in classifier. In mapping function $T$, $l$ is set to $0.5$. We update $\lambda$ according to \cite{long2018conditional}. For parameters of each baseline, we all follow the original settings. We implement all methods with default parameters by PyTorch. The code of CLARINET is available at \httpsurl{github.com/Yiyang98/BFUDA}.

\subsection{Results on BFUDA Tasks}
Table \ref{tab:ccom} reports the target-domain accuracy of $5$ methods on $6$ BFUDA tasks. As shown, our CLARINET performs best on each task and the average accuracy of CLARINET is significantly higher than those of baselines. Compared with GAC method, CLARINET successfully transfers knowledge from complementary-label source data to unlabeled target data. Since CDAN has shown much better adaptation performance than DANN and DAN \cite{long2018conditional}, GAC+CDAN should outperform other two-step methods on each task. However, on the task $U$$\rightarrow$$ M$, the accuracy of GAC+CDAN is much lower than that of GAC+DANN. This abnormal phenomenon shows that the noise contained in pseudo-label source data significantly reduces transferability of existing UDA methods. Namely, we cannot obtain the reliable adaptation performance by using two-step BFUDA approach.


Figure \ref{fig:ccom} shows the target-domain accuracy of two-step methods and CLARINET on $6$ BFUDA tasks when increasing epochs. It is clear to see that the accuracy of CLARINET gradually and steadily increases and eventually converges, achieving the best accuracy on each task. The accuracy of GAC+CDAN always reaches plateau quickly. For GAC+DANN, its accuracy is unstable and significantly drops after certain epochs on $M$$\rightarrow$$ U$, $M$$\rightarrow$$ m$ tasks. While the accuracy of GAC+DAN is relatively stable but not satisfactory.

\subsection{Ablation Study}

Finally, we conduct experiments to show the contributions of components in CLARINET. We consider following baselines: 

\begin{itemize}
\item C w/ $L_{CE}$: train \underline{C}LARINET by Algorithm~\ref{alg:algorithm}, while {replacing} $\overline{L}_s(G,F,\overline{D}_s)$ by \underline{c}ross-\underline{e}ntropy loss.
\item C w/o $T$: train \underline{C}LARINET by Algorithm~\ref{alg:algorithm}, \underline{without} mapping function $\underline{T}$.
\end{itemize}

C w/ $L_{CE}$ uses the cross-entropy loss to replace the complementary-label loss. The target-domain accuracy of C w/ $L_{CE}$ will show if UDA methods can address the BFUDA problem. Comparing CLARINET with C w/o $T$ reveals whether the mapping function $T$ takes effect. 
As shown in Table~\ref{tab:ablation}, the target-domain accuracy of C w/ $L_{CE}$ is much lower than that of other methods. Namely UDA methods cannot handle BFUDA tasks directly. Although C w/o $T$ achieves better accuracy than two-step methods, its accuracy is still worse than CLARINET's. Thus, the mapping function $T$ can help improve the adaptation performance under BFUDA.  


\section{Conclusion}

This paper presents a new problem setting for the domain adaptation field, called {\em budget-friendly unsupervised domain adaptation} (BFUDA), which exploits economical complementary-label source data instead of expensive true-label source data. Since existing UDA methods cannot address BFUDA problem, we propose a novel one-step BFUDA approach, called \emph{\underline{c}omplementary \underline{l}abel advers\underline{ari}al \underline{net}work} (CLARINET). Experiments conducted on $6$ BFUDA tasks confirm that CLARINET effectively achieves distributional adaptation from complementary-label source data to unlabeled target data and outperforms competitive baselines. 
\section*{Acknowledgements}

The work presented in this paper was supported by the Australian Research Council (ARC) under FL190100149 and DP170101632. We would like to thank the anonymous reviewers for their thoughtful comments. The first author particularly thanks the support by UTS-CAI during her visit.

\bibliographystyle{named}
\bibliography{ijcai20}

\newpage
\newpage

\includepdf[pages=1]{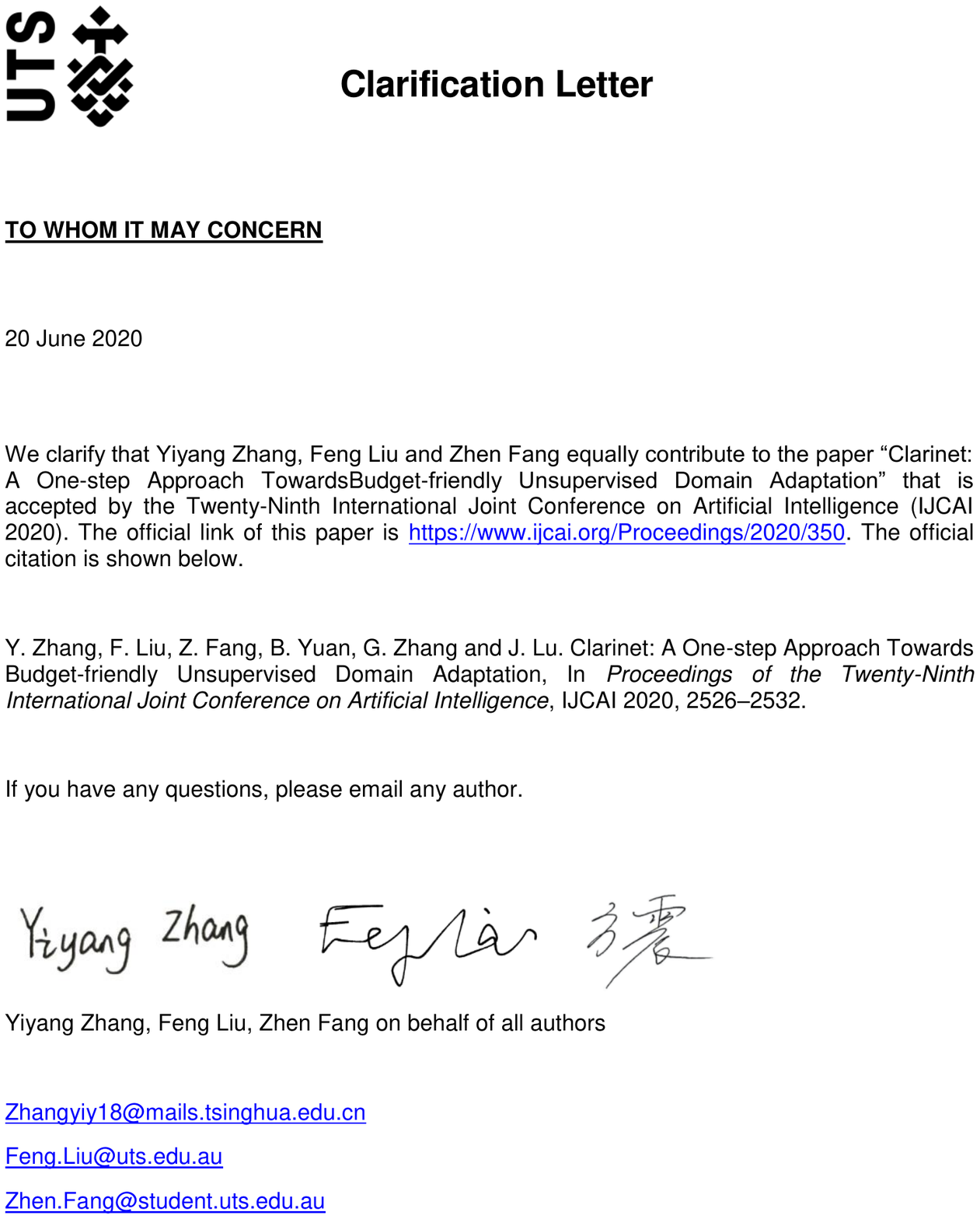}

\end{document}